\documentclass{styles/svproc}
\usepackage{url}
\usepackage{cite}
\usepackage{amsmath,amssymb,amsfonts}
\usepackage{algorithmic}
\usepackage{graphicx}
\usepackage{textcomp}
\usepackage{xcolor}
\usepackage{wrapfig}
\usepackage{multirow}
\usepackage{subcaption}
\usepackage{subfig}
\usepackage{array}
\usepackage{tabularx}
\usepackage{marvosym}

\begin{document}
\mainmatter              % start of a contribution

\title{Learning-based Traversability Costmap for Autonomous Off-road Navigation}
\titlerunning{Learning-based Traversability Costmap}  % abbreviated title (for running head)
\author{Qiumin Zhu\inst{1} \and Zhen Sun\inst{1} \and
Songpengcheng Xia\inst{1} \and Guoqing Liu\inst{1} \and Kehui Ma\inst{1} \and Ling Pei\inst{1}\textsuperscript{\Letter} \and
Zheng Gong\inst{2}\textsuperscript{\Letter} \and Cheng Jin\inst{3}}
\authorrunning{Qiumin Zhu et al.} % abbreviated author list (for running head)
\tocauthor{Qiumin Zhu, Zhen Sun,
Songpengcheng Xia, Guoqing Liu, Kehui Ma, Ling Pei and 
Zheng Gong}
\institute{Shanghai Key Laboratory of Navigation and Location Based Services,\\ Shanghai Jiao Tong University, Shanghai, China,\\
\email{ling.pei@sjtu.edu.cn}
\and
China Academy of Information and Communications Technology, Beijing, China, \\
\email{gongzheng1@caict.ac.cn}
\and
Key Laboratory of Smart Earth, Beijing, China, \\
\email{jinchengno1@163.com}}
\renewcommand{\thefootnote}{}
\footnotetext{\Letter Corresponding author}
\maketitle              % typeset the title of the contribution
\begin{abstract}
Traversability estimation in off-road terrains is an essential procedure for autonomous navigation. However, creating reliable labels for complex interactions between the robot and the surface is still a challenging problem in learning-based costmap generation. To address this, we propose a method that predicts traversability costmaps by leveraging both visual and geometric information of the environment. To quantify the surface properties like roughness and bumpiness, we introduce a novel way of risk-aware labelling with proprioceptive information for network training. We validate our method in costmap prediction and navigation tasks for complex off-road scenarios. Our results demonstrate that our costmap prediction method excels in terms of average accuracy and MSE. The navigation results indicate that using our learned costmaps leads to safer and smoother driving, outperforming previous methods in terms of the highest success rate, lowest normalized trajectory length, lowest time cost, and highest mean stability across two scenarios.
% We would like to encourage you to list your keywords within
% the abstract section using the \keywords{...} command.
\end{abstract}
\section{Introduction}

Autonomous navigation in off-road environments is a critical problem for unmanned ground vehicles. Robots are utilized in wilderness, forests, mines and other complex terrains for tasks like agriculture, mining, planetary exploration, surveillance and so on. To ensure safe driving, it is crucial to analyze the traversability of these terrains and construct a costmap for navigation.

Various works focus on this issue and have contributed to the progress of research. Initially, the traversability estimation was regarded as a binary classification problem to differentiate between traversable and untraversable terrains. Currently, it can be viewed as the multiple classes categorization relevant to the level of traverse difficulty \cite{c1,c2} or the regression to assign a continuous traversability value \cite{tc2}. Despite remarkable efforts on evaluating traversability, there are two remaining challenges for the research community.

One challenge is the representation of the terrain characteristics. The statistics of geometric properties is a popular approach, building a height grid map to calculate step height, roughness and slope as traversability score \cite{height1,height2}. The appearance-based method redirects the problem to image processing and classification. The terrain is categorized by texture and color of the image and recently semantic segmentation has become a useful tool for traversability estimation \cite{app2,s1}. While the geometry and appearance information can assess traversability separately, there are conditions with the same geometric properties or appearance but different traversability such as dry mud and wet mud, or low grass and high grass. We consider both geometric and visual characteristics of environments to evaluate traversability comprehensively. While such information can represent different terrains, the actual impact of the ground on the vehicle that determines the traversability is still unknown.

Therefore, the other challenge concerns the definition of traversability cost labels. Although a variety of works propose learning-based costmap approaches, they predict the cost ignoring the nuance of the robot's interactions with different terrains. Some simply divide the ground into traversable and untraversable whether the vehicle can reach or not \cite{lc3}, while others represent slip as the velocity error \cite{lc4}. To capture roughness and bumpiness that the vehicle experiences, proprioceptive sensors like Inertial Measurement Units (IMUs), are useful for sensing these characteristics. It is easy for IMUs to build on the ground vehicle and capture the state \cite{360}. The signals reflect the vehicle's vibrations, movements, and changes in feelings as traversability costs. We take the properties of IMU data and robotic risk into consideration, handling the IMU linear acceleration in the z-axis to generate continuous traversability values as learning targets.

In brief, we propose a learning-based method to predict traversability costmaps that present risk-aware influence of the terrains with respect to the robot navigation. Exteroceptive information including semantic and geometric features and interoceptive information as the robot velocity are the inputs for learning a continuous cost which is supervised by processed IMU data. We combine a Convolutional Neural Network (CNN) backbone to extract features from exteroceptive information, a neural network to process interoceptive information and a Long Short-Term Memory (LSTM) to handle the concatenate features as our learning architecture.

The main contributions of this research are as follows:
\begin{itemize}
    \item[$\bullet$] We propose a learning-based framework that predicts a continuous traversability costmap from a fusion of semantic and geometric information, as well as linear and angular velocity, using an LSTM to process sequences of these features.
    \item[$\bullet$] To reflect the reaction and risk of driving on different terrains, we present a novel traversability cost label that is derived from IMU data for training the network. 
    \item[$\bullet$] We validate our method on both costmap prediction and navigation tasks in two off-road scenarios, outperforming in terms of accuracy, MSE, success rate, time cost and stability of driving.
\end{itemize}

\section{Methods}

\subsection{Overview}
\begin{figure*}[t]
    \centering
    \includegraphics[width=\textwidth]{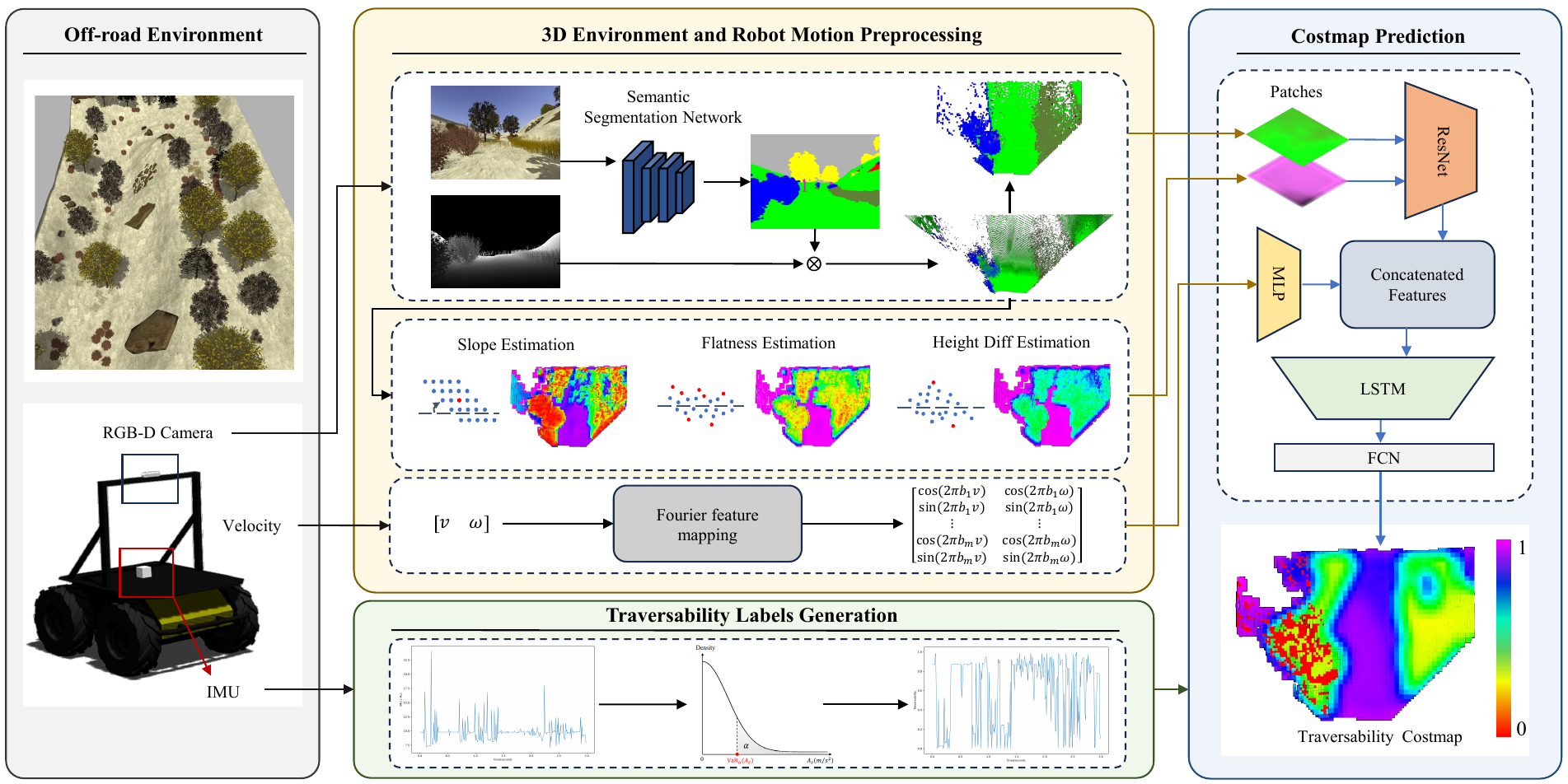}
    \caption{Overview of the traversability costmap prediction system. Our method includes three modules: 1) traversability labels generation; 2) 3D environment and robot motion preprocessing; and 3) costmap prediction. We use IMU data to compute risk-aware traversability labels. 3D environment information and velocity are preprocessed to be inputs of the network. A continuous traversability costmap is produced by the network for navigation in off-road terrains.}
    \label{fig:overview}
\end{figure*}

We propose a learning-based framework to assess terrain traversability and obtain a costmap for navigation in off-road environment. It takes RGB-D point cloud and robot velocity as inputs of a neural network, and outputs a robot-centric traversability costmap. The overview of the framework is illustrated in Fig. \ref{fig:overview}. The costmap predicted by the network can be used by the path planning algorithm to realize autonomous navigation.

The framework is decomposed into three main modules: 1) traversability labels generation; 2) 3D environment and robot motion preprocessing; and 3) costmap prediction. The labels generation step calculates traversability cost from proprioception. The preprocessing module extracts semantic and geometric data from RGB-D images and point clouds, and represent robot velocity, as the inputs of the network. With the inputs and labels, a neural network is trained to predict traversability cost.

\subsection{Traversability Labels Generation}
\begin{figure}[th]
  \centering
  \includegraphics[width=0.8\textwidth]{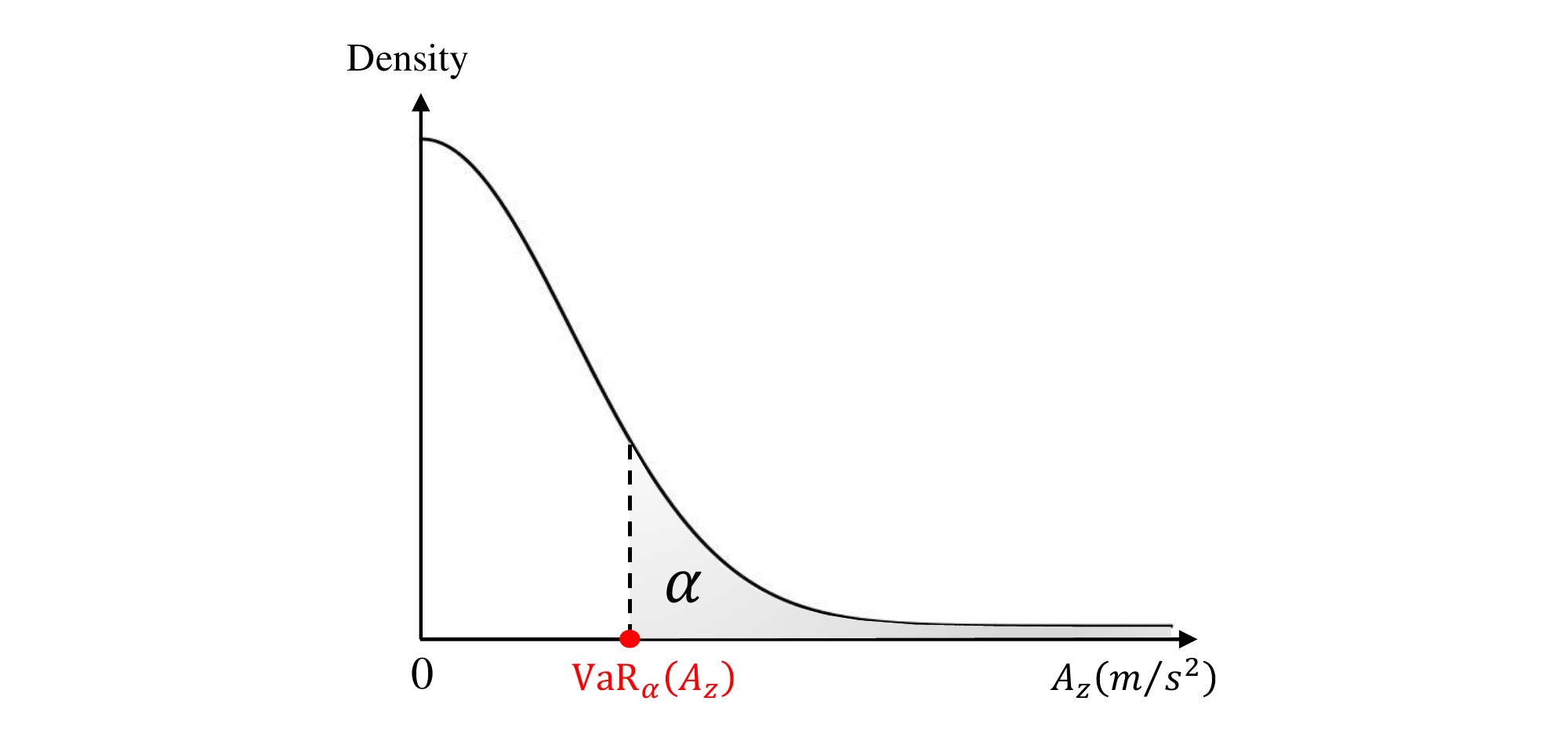}
  \caption{This work defines IMU data as value at risk (VaR) and traversability labels as risk level $\alpha$.}
  \label{fig:var}
\end{figure}

To learn a continuous and normalized traversability cost, we use linear acceleration in the z-axis to describe the interactions between the robot and the ground \cite{imu3,imu4,imu5}. The z-axis linear acceleration, generally understood as the force acting on the z-axis of the robot, reflects the roughness and bumpiness of the terrain. As shown in previous work \cite{imu_gaussian,imu_gaussian_2}, the IMU linear acceleration measurements follow normal or Gaussian distributions in stationary conditions. With the physical properties of the z-axis linear acceleration, the mean of the distribution is approximately the value of gravitational acceleration. We adopt Value at Risk (VaR), which is used as assessment of robot risk \cite{risk}, to quantify traversability cost. Since values that deviate equally from the mean on both sides can be considered to have the same cost, the distribution $X\sim N\left(\mu, \sigma^2\right)$ can be transformed into a half-normal distribution by $\left|\frac{X - \mu}{\sigma}\right|$ and $\text{VaR}_{\alpha}(A_{z})$ at level ${\alpha}$ is simply the $(1-{\alpha})$-quantile, shown in Fig. \ref{fig:var}:
\begin{equation}
\operatorname{VaR}_\alpha(A_{z}):=\min\{a_{z}\mid\mathbb{P}[A_{z}>a_{z}]\leq\alpha\}
\end{equation}
We take each processed IMU z-axis linear acceleration value as $\text{VaR}_{\alpha}(A_{z})$ and calculate its risk level ${\alpha}\in\left[0,1\right]$ as the traversability cost label, where 1 means lowest cost and 0 means highest cost. The data recorded during a steady motion is used to derive the distribution.

\subsection{3D Environment and Robot Motion Preprocessing}
To acquire inputs of the learning network, we represent terrain characteristics about the environment as visual and geometric information and parameterize the robot velocity to high-dimensional. We use Fast-SCNN \cite{fast-scnn} to predict the semantic segmentation of a RGB image from the RGB-D camera and generate colored point cloud based on the semantic and depth images. A local grid map with a resolution of 0.1 meter is built from the point cloud and each grid cell contains RGB and geometric values, regardless of points 2 meters above the vehicle that have no contact.

To calculate the geometric estimation including slope, flatness and height difference, we first construct an octree of the local point clouds to expedite the retrieval of points in each location. The slope $s$ in each grid cell is represented by the angle between the z-axis of the vehicle coordinate frame and the surface normal of a square area with a width of 0.5 meter centered around the grid cell:
\begin{equation}
    s=\frac{180\arccos{\left(\mathbf{n}\cdot \mathbf{e_z}\right)}}{\pi}
\end{equation}
where $\mathbf{n}$ is the unit normal vector calculated with PCA \cite{hc2} and $\mathbf{e_z}$ is vector $\left[0,0,1\right]^\top$. The range of the slope value is from 0 to 90 degrees.
The flatness $f$ is calculated by the vertical height of points in arbitrary grid cell:
\begin{equation}
    f = \sqrt{\frac{\sum_{j=1}^N\left[\mathbf{n}\cdot(\mathbf{p_j}-\bar{\mathbf{p}})\right]^2}{N+1}}
\end{equation}
where $\bar{\mathbf{p}}$ is the 3D centroid of the grid cell, $\mathbf{p_j}=[x,y,z]^\top$ is the position of points in the grid cell and $N$ is the number of these points. The height difference $h$ is computed as the max deviation between the vertical height of points:
\begin{equation}
    h = \max\left[\mathbf{n}\cdot(\mathbf{p_i}-\bar{\mathbf{p}})\right]-\min\left[\mathbf{n}\cdot(\mathbf{p_j}-\bar{\mathbf{p}})\right], i,j\in[1,N]
\end{equation}

When driving at high speeds or making sharp turns, the vehicle can sense significant vibrations caused by bumpy terrains. We consider the influence of the velocity and process it as an input of our network. To match the dimension of the local grid map, we use Fourier features \cite{imu5} to parameterize the velocity to a higher dimensional vector $\lambda(v)$:
\begin{equation}
    \lambda(v) = \left[
    \begin{array}{cc}
         \cos \left(2 \pi b_1 v\right) & \cos \left(2 \pi b_1 \omega\right) \\
         \sin \left(2 \pi b_1 v\right) & \sin \left(2 \pi b_1 \omega\right) \\
         \vdots  & \vdots \\
         \cos \left(2 \pi b_m v\right) & \cos \left(2 \pi b_m \omega\right) \\
         \sin \left(2 \pi b_m v\right) & \sin \left(2 \pi b_m \omega\right)
    \end{array}\right]
\end{equation}
where $v$ is the norm of linear velocity in the x-axis and y-axis, $\omega$ is z-axis angular velocity, $b_i \sim \mathcal{N}\left(0, \sigma^2\right)$ are sampled from a Gaussian distribution and $m$ corresponds to the scale of the local map patches.

\subsection{Costmap Prediction}
The pipeline of our costmap prediction is similar to \cite{imu5}. We first extract local map patches from robot trajectories, then predict traversability costs for these patches by training a CNN-LSTM network, and finally generate a costmap for navigation with the trained network.

 We collect all environment data to construct a global map, then locate and extract $1\times1$ meter patches in the global map with a set of robot odometry information sampled per 0.1 second. The linear and angular velocity is also recorded at these positions. We compute the average risk level using five consecutive frames of IMU linear acceleration data at each position as the ground-truth label for traversability cost.

We train a network to predict a continuous value from 0 to 1 as the traversability cost with inputs of semantic and geometric map patches, as well as parameterized velocity. We concatenate these features extracted by ResNet18 \cite{resnet} and MLP, then pass them through an LSTM \cite{lstm} to handle sequences of data since the point cloud patches are sampled along the driven trajectory of a global map. We use the same loss and optimizer as \cite{imu5}.

To produce real-time costmaps during navigation, we take $10\times10$ meter point cloud local map generated from the current RGB image and depth image. The $1\times1$ meter patches are downsampled from the local map per 0.2 meter. With these patches and velocity, the network predicts traversability costs and the final value of each $0.1\times0.1$ meter cell is the average of traversability costs of all patches that cover this cell. The costmap maintains the shape of the local map input by checking the cell with no point and removing it from the output. While the network only learns the information of the environment that the vehicle can traverse, it ignores the obstacle data of where is unreachable. We record the semantic classes that the robot cannot travel during the dataset collection and assign a value of 0 to the cells containing these classes.

\section{Experiment and Results}

\subsection{Simulation Setup}
\begin{figure}[htbp]
\centering 
\begin{subfigure}[b]{0.48\linewidth} 
    \centering
    \includegraphics[width=\linewidth]{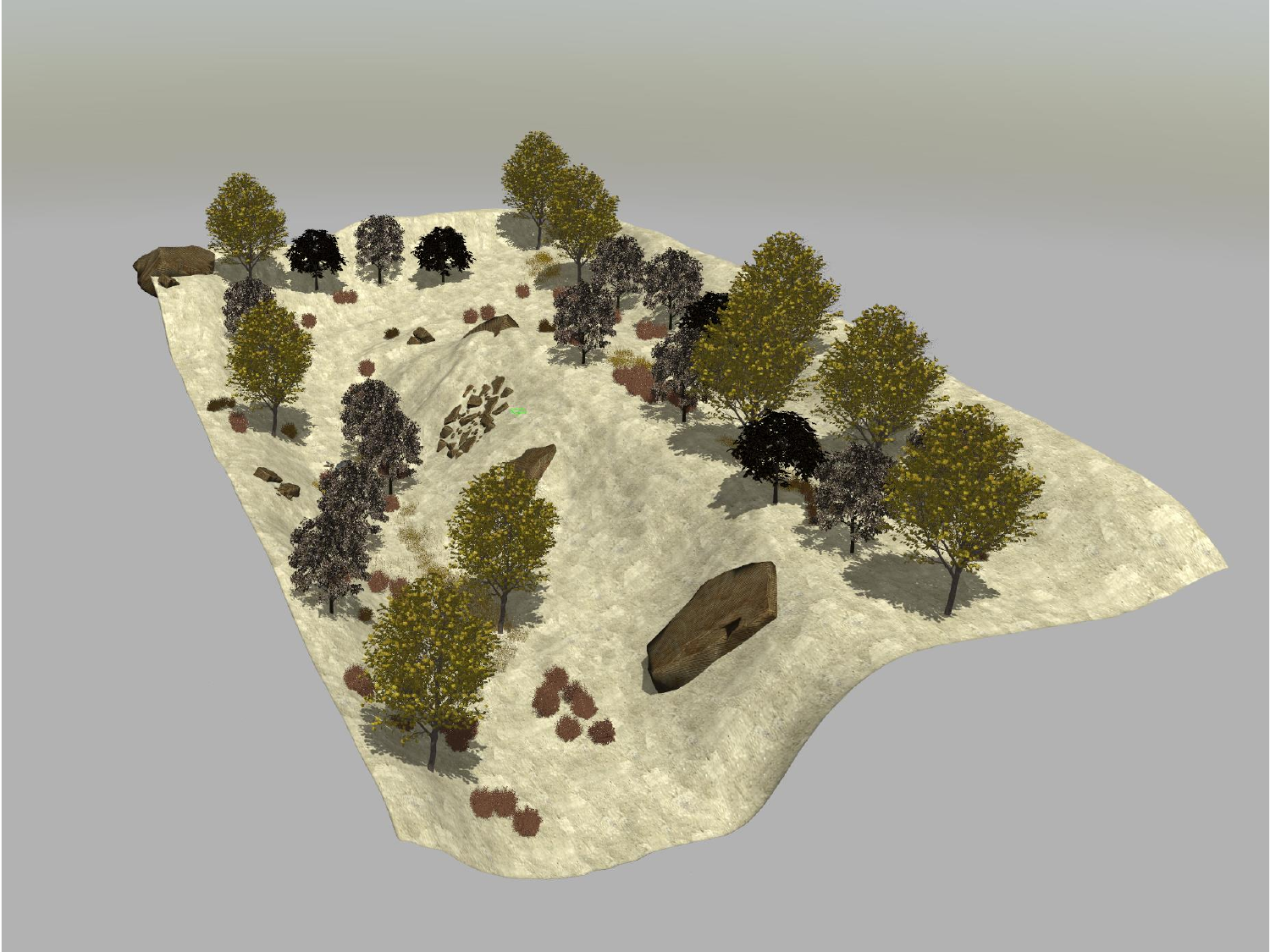} 
    \subcaption{hill} 
    \label{fig:hill_sim}
\end{subfigure}
\begin{subfigure}[b]{0.48\linewidth} 
    \centering
    \includegraphics[width=\linewidth]{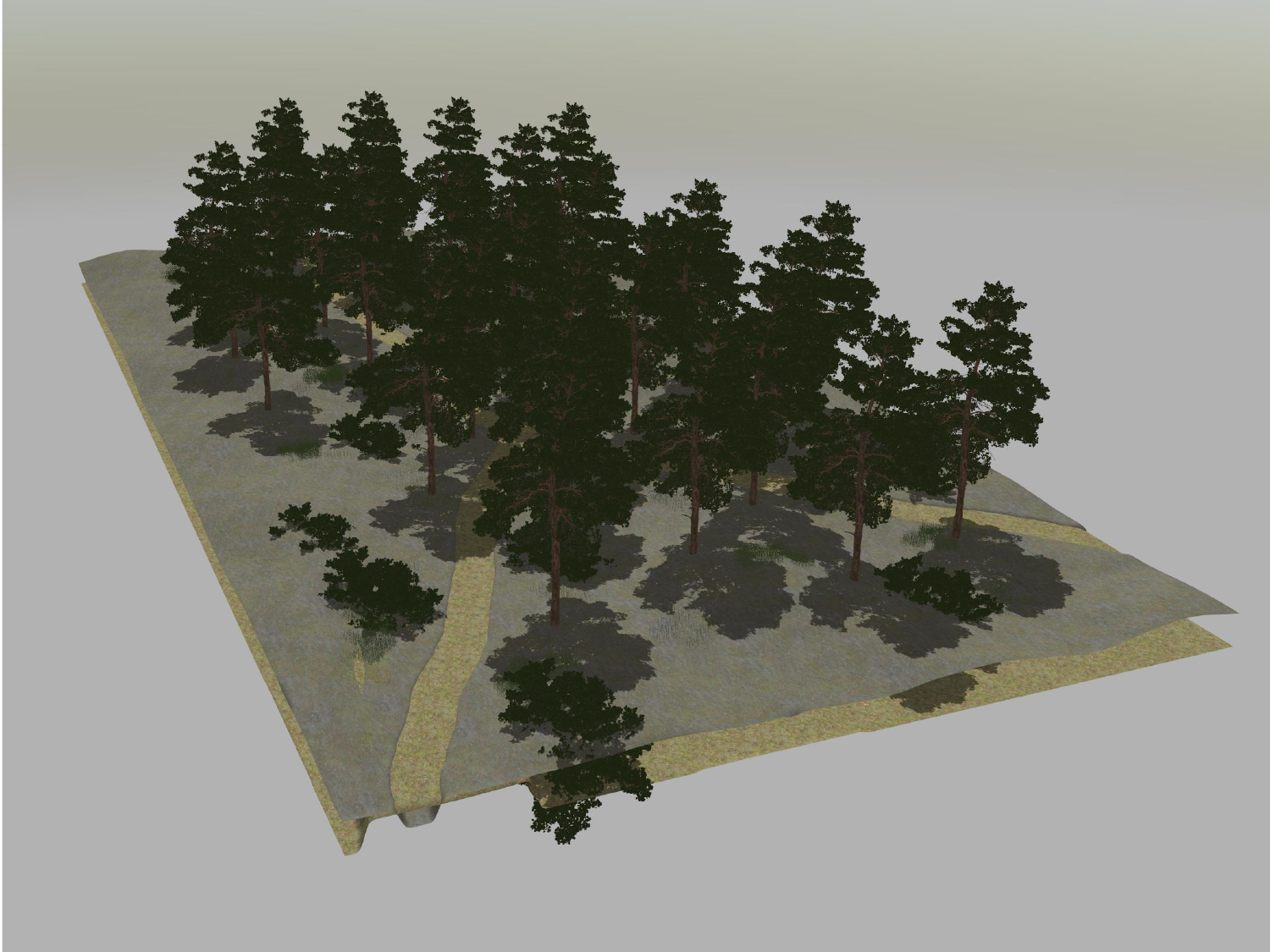}
    \subcaption{forest} 
    \label{fig:forest_sim}
\end{subfigure}
\caption{Two Gazebo simulation scenarios for experiments. (a) is a steep hill and (b) is a dense forest.}
\label{fig:simulation}
\end{figure}

We use the natural environments in Gazebo simulation \cite{gazebo} to train and test our framework. A HUSKY robot is equipped with a RGB-D camera and an IMU. A Gazebo plugin is used to obtain the ground-truth odometry of the vehicle. Two scenarios, as shown in Fig. \ref{fig:simulation}, are employed for experimentation: i\textbf{)} the rugged hillside, which contains steep slopes with high grass, bushes, rock and trees; ii\textbf{)} the dense forest, which consists of high grass, stones, bushes, trees and fallen trunks. In order to enhance the realism of the scene, we adjust the physical properties of high grass so that the vehicle can pass through but with resistance. We also modify the environments to be more challenging for navigation by adding grass, rocks and bushes in different places. With the Robot Operating System (ROS), the data is recorded and the vehicle is controlled. The algorithm and the simulation are processed by a desktop computer with Intel i5-12490F CPU, 32GB RAM, and Nvidia RTX 3070 GPU.

\subsection{Training Data}
We collect data in the simulation environments by following different paths for our network training. We use RGB and depth images in the dataset to obtain dense colored point clouds and the IMU to obtain the ground-truth traversability costs. The odometry data including position, orientation and velocity is provided by the plugin. The ratio of low to high cost frames is 2:1 in the hillside scenario for its complexity and roughness, while it is 5:1 in the forest scenario because of the flatness. In total, we generate 4K training frames, 0.5K validation frames and 0.5K test frames for our experiment.

\subsection{Costmap Evaluation}
\begin{figure}[th]
  \centering
  \includegraphics[width=1\textwidth]{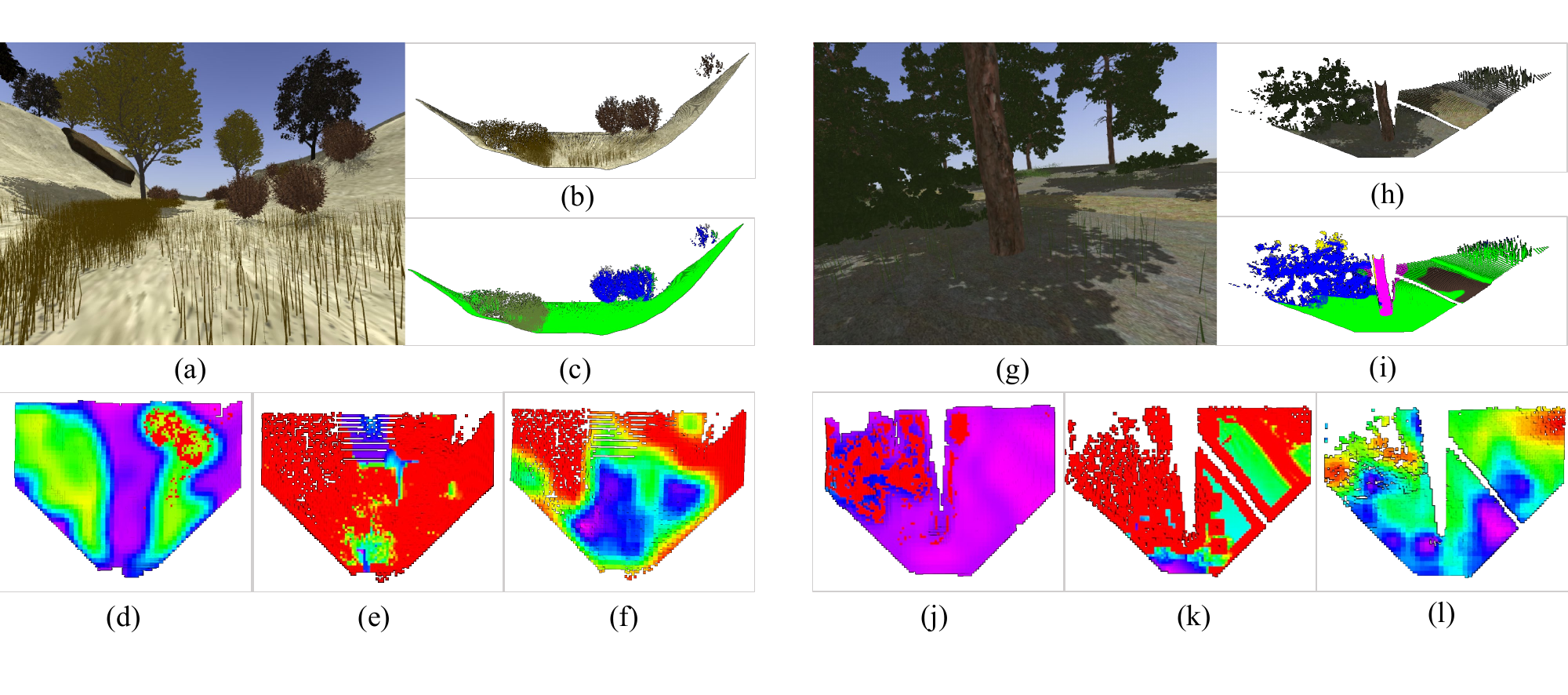}
  \caption{Costmap prediction comparison. (a) and (g) are the robot's front-facing view. (b) and (h) show the RGB-D point clouds. (c) and (i) show the semantic point clouds. (d) and (j) are costmaps predicted by our method, (e) and (k) are TNS's, (f) and (l) are HDIF's. Rainbow colors visualize traversability cost, where red is 0 and purple is 1.}
  \label{fig:comparison}
\end{figure}

We compare the performance with two appearance and geometry based baselines to evaluate the quality of our method. TNS \cite{hc2} uses a non-learning approach that calculates the traversability score based on terrain classes and geometric traversability including slope and step height. HDIF \cite{imu5} characterizes roughness as the 1-30 Hz bandpower of the IMU z-axis linear acceleration and trains a network to learn the traversability costs. This network is trained on our dataset. Fig. \ref{fig:comparison} shows the costmaps predicted by these three methods at the same location. Although these methods output a continuous value between 0 and 1, we simplify the comparison to avoid any biases. We use the ground-truth semantic environments to populate a traversability grid map by converting the labels of the point cloud to either 0 or 1. Traversable regions like ground, high grass and trail are set to be 1, and other regions like trunk, bush and rock are set to be 0.

The vehicle travels along the paths both in realistic and semantic environments. The global costmaps are generated by the baselines and our method with the collected sensor data and evaluated with four different metrics, similar to \cite{hc2}. The metrics are described as follows:

\noindent\textbf{Trav. Accuracy:} The accuracy of traversable regions.

\noindent\textbf{All Accuracy:} The accuracy over all grids of the map.

\noindent\textbf{ROC (Receiver Operation Curve):} We set the cost exceeding 0.5 to 1, and the rest to 0 as a binary classification and use ROC to indicate the performance through true positive and false positive rates.

\noindent\textbf{MSE (Mean Squared Error):} We calculate the average distance between the continuous predictions and the ground truth labels to estimate the quality of the predictions.

\begin{table}[h]
\centering
\caption{Results of Costmap Comparisons}
\label{table_costmap}
\begin{tabular}{|c|c|c|c|c|c|}
\hline
Scene & Method & Trav. Acc $\uparrow$ & aAcc $\uparrow$ & AUC $\uparrow$ & MSE $\downarrow$ \\
\hline
\multirow{4}{*}{hill} & TNS\cite{hc2}  & 86.34 & 86.12 & 0.894 & \textbf{0.117} \\
 & HDIF\cite{imu5} & 14.80  & 55.95 & 0.695 & 0.662 \\
 & ours & \textbf{92.39}  & \textbf{91.52} & \textbf{0.897} & 0.149 \\
\hline
\multirow{4}{*}{forest} & TNS\cite{hc2} & 40.73  & 42.07 & \textbf{0.920} & 0.405 \\
 & HDIF\cite{imu5} & 2.56  & 5.00 & 0.795 & 0.463 \\
 & ours & \textbf{99.07} & \textbf{97.48} & 0.914 & \textbf{0.0581} \\
\hline
\end{tabular}
\end{table}

\begin{figure}[htbp]
\centering 
\begin{subfigure}[b]{0.49\linewidth} 
    \centering
    \includegraphics[width=\linewidth]{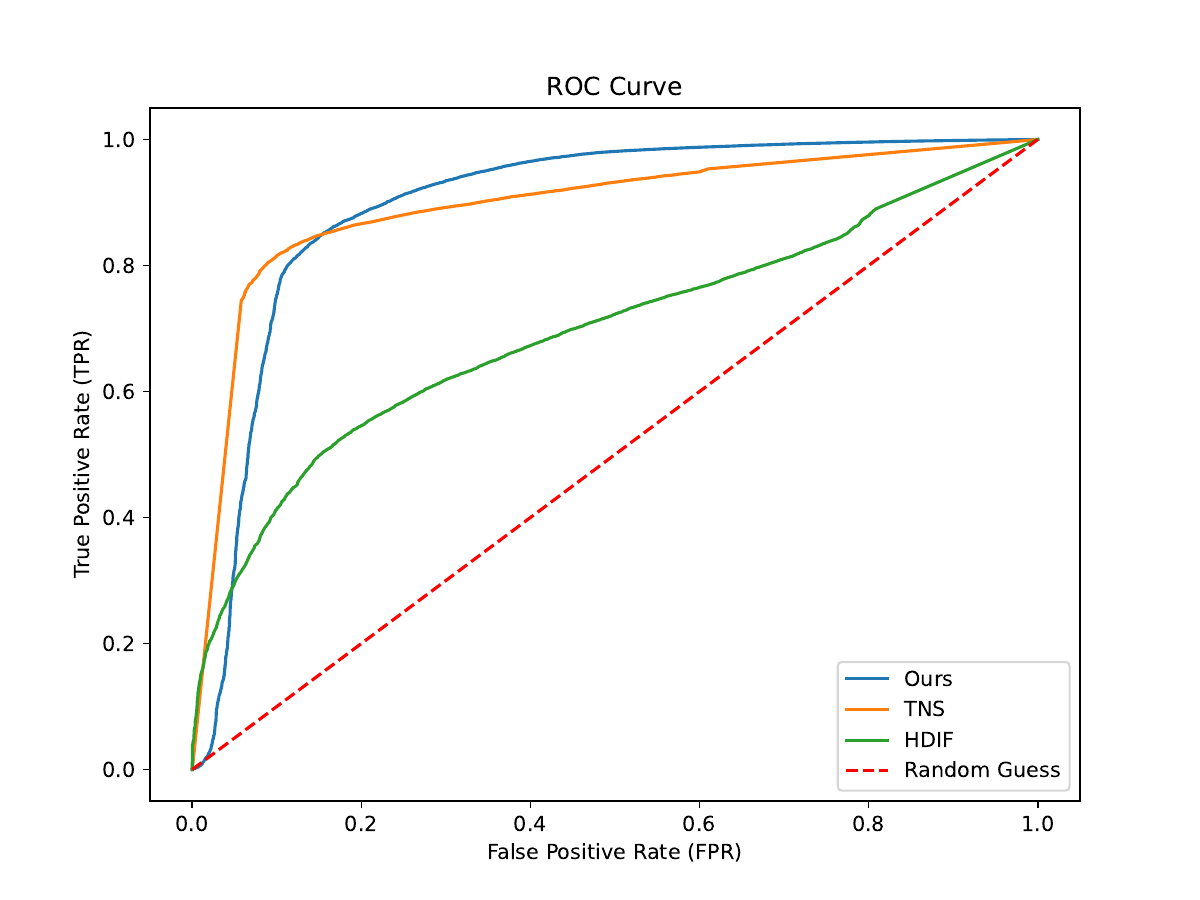} 
    \subcaption{hill} 
    \label{fig:hill_roc}
\end{subfigure}
\begin{subfigure}[b]{0.49\linewidth} 
    \centering
    \includegraphics[width=\linewidth]{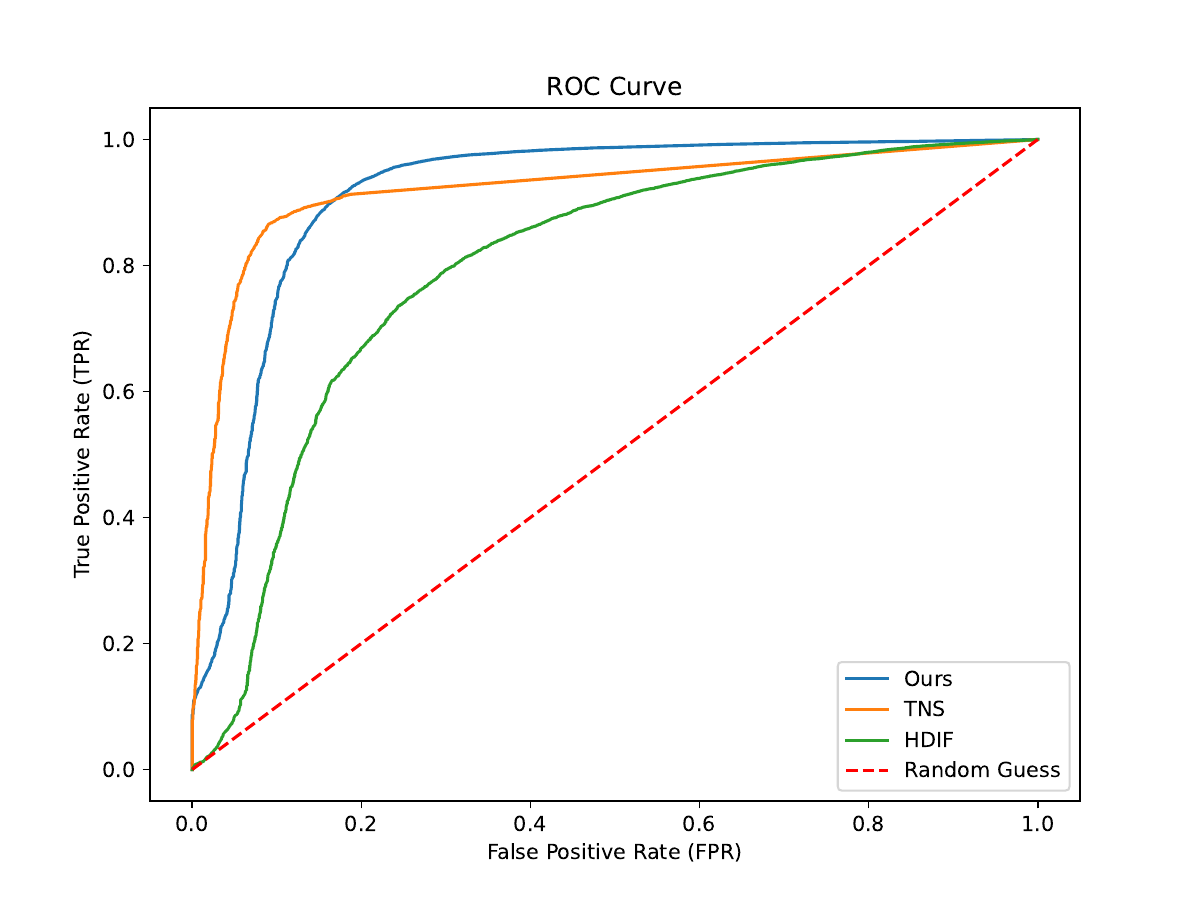}
    \subcaption{forest} 
    \label{fig:foreset_roc}
\end{subfigure}
\caption{ROC plots. We plot ROCs on two baselines and our method. (a) is the result of the hill scenario and (b) is the result of the forest scenario.}
\label{fig:roc}
\end{figure}

Table \ref{table_costmap} shows the comparison of costmap prediction performance in two scenarios. Our method has better performance on accuracy and MSE. Fig. \ref{fig:roc} indicates that our method performs similarly to TNS on AUC, both superior to HDIF. TNS divides the semantic classes into traversable and untraversable terrains in advance, hence it has a little advantage over ours. The result of the costmap comparisons verifies that our definition of traversability cost is reasonable and our costmap prediction is accurate.

\subsection{Navigation Evaluation}
We validate our costmap for autonomous navigation in off-road environments and compare the performance with the two baselines. A real-time costmap is generated by the current frame of RGB images, depth images and odometry data. We use the output point cloud of the costmap for local path planning by combining it with a collision-free path planning algorithm \cite{falco}.

We design 8 trials in each scenario and use the following metrics \cite{imu1} to evaluate the performance of the navigation:

\noindent\textbf{Success Rate ($R_{success}$):} The proportion of the robot achieving the goal.

\noindent\textbf{Normalized Trajectory Length ($\bar{L}$):} The trajectory length normalized by the Euclidean distance between the start and the goal for all successful trials.

\noindent\textbf{Relative Time Cost ($t_{rel}$):} The ratio of the time cost in the same successful trials of other methods to ours.

\noindent\textbf{Mean Stability ($\bar{S}$):} The traversability cost calculated by the IMU data. We compute the mean traversability cost of all frames of IMU z-axis linear acceleration as the stability of the robot in trials.

\begin{table}[h]
\centering
\caption{Results of Navigation Comparisons}
\label{table_navigation}
\begin{tabular}{|c|c|c|c|c|c|}
\hline
\rule{0pt}{9pt}
Scene & Method &  $R_{success}$ (\%) $\uparrow$ & $\bar{L}$ ($\to1$)  & $t_{rel}$ $\downarrow$  & $\bar{S}$ $\uparrow$\\
\hline
\multirow{4}{*}{hill} & TNS\cite{hc2} & 47.46 & 1.505 & 1.648 & 0.727 \\
 & HDIF\cite{imu5} & 29.28 & 2.647 & 3.658 & 0.712 \\
 & ours & \textbf{100} & \textbf{1.103} & \textbf{1.000} & \textbf{0.799} \\
\hline
\multirow{4}{*}{forest} & TNS\cite{hc2} & 40.48 & 1.407 & 1.890 &  0.758\\
 & HDIF\cite{imu5} & 69.67 & \textbf{1.090} & 1.047 & 0.799 \\
 & ours & \textbf{96.82} & 1.098 & \textbf{1.000} & \textbf{0.882} \\
\hline
\end{tabular}
\end{table}

\begin{figure}[htbp]
\centering 
\begin{subfigure}[b]{0.61\textwidth} 
    \centering
    \includegraphics[width=\textwidth]{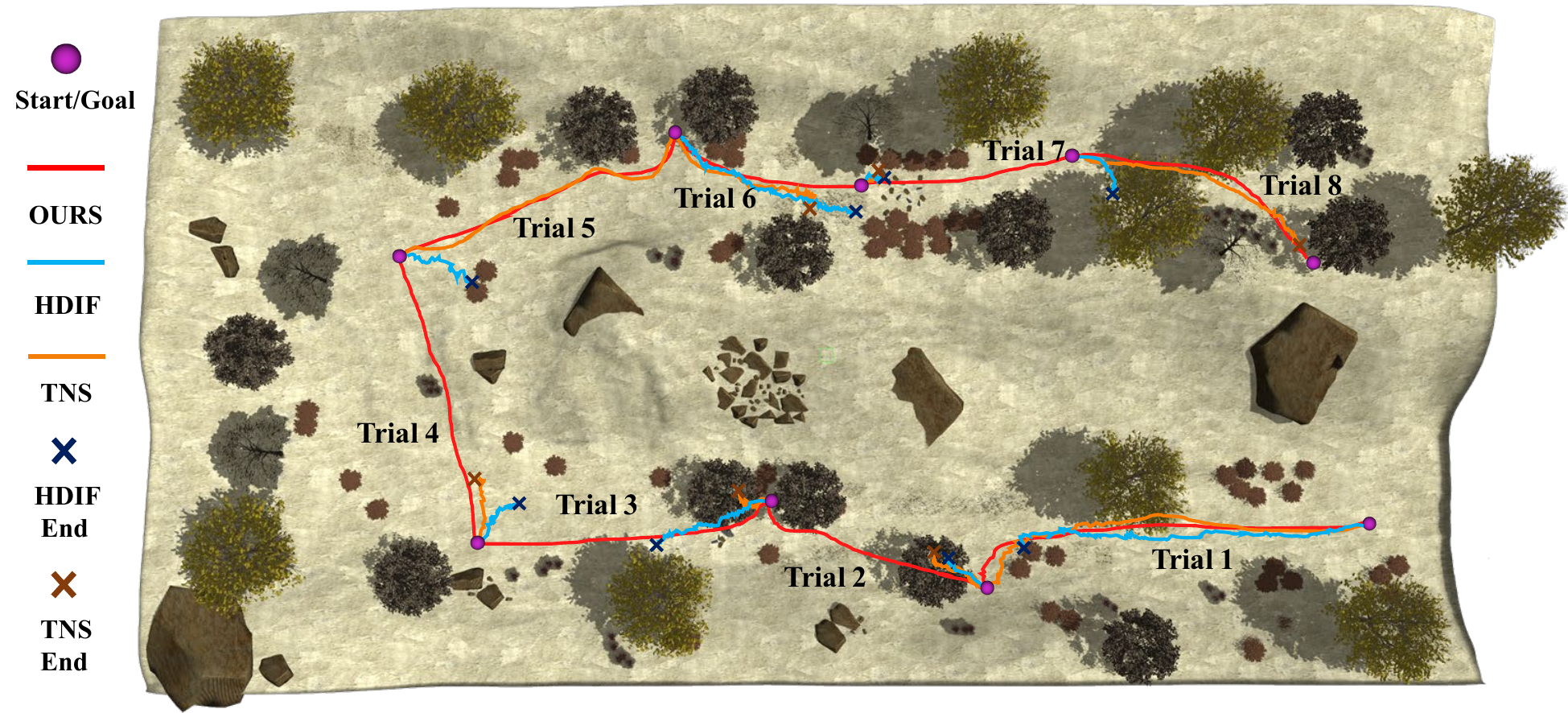} 
    \subcaption{hill} 
    \label{fig:hill_nav}
\end{subfigure}
\begin{subfigure}[b]{0.61\textwidth} 
    \centering
    \includegraphics[width=\textwidth]{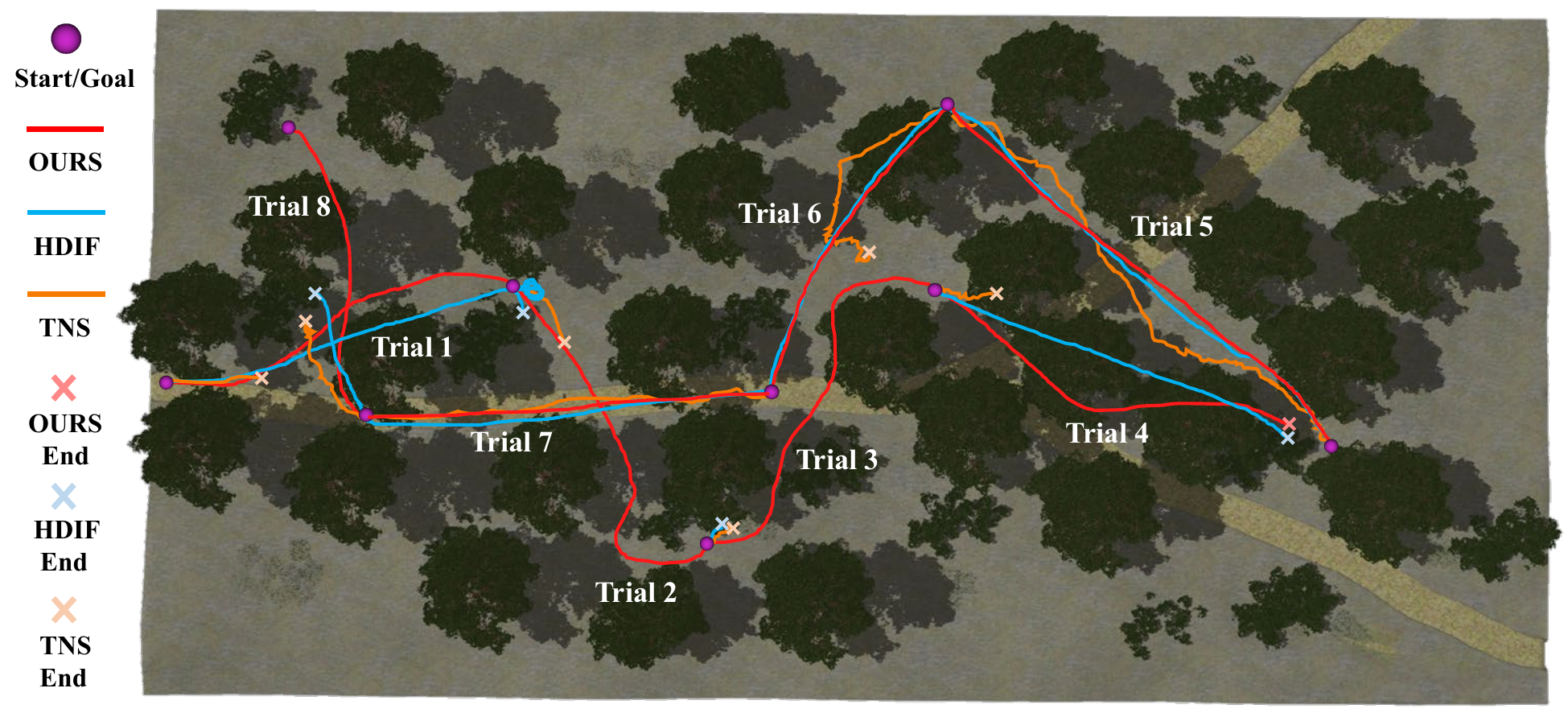}
    \subcaption{forest} 
    \label{fig:foreset_nav}
\end{subfigure}
\caption{Navigation experiment. The purple points are the starts and the goals of trials. (a) is the result of the hill scenario and (b) is forest. Red trajectories are ours, orange ones are TNS's, and blue ones are HDIF's. }
\label{fig:nav}
\end{figure}

Table \ref{table_navigation} shows the comparison of navigation performance in two scenarios. Our method outperforms the other methods in terms of all four metrics. HDIF takes shorter trajectories in the forest scene since it travels directly towards the goal without considering the obstacles and its successful trajectories for evaluation are less than ours. A higher success rate of our method shows that it is effective even in different and complex situations while the other methods have failed. The shorter normalized trajectory length of our method indicates that it is more efficient and precise by minimizing unnecessary movements and simultaneously avoiding collision. The less time cost and higher stability in the condition of the same preset speed demonstrate that our method ensures a faster but safer and more stable navigation. The trajectories of the navigation experiment are illustrated in Fig. \ref{fig:nav}.

\subsection{Further Analysis}
We use angular velocity as the inputs and an LSTM in the network. To validate whether they improve the performance of our network, we make an ablation study. We use the same dataset to train three networks: ours, ours without angular velocity $\omega$ and ours without LSTM. We compare the three best learned models in the validation set and test set. Table \ref{table_loss} shows that angular velocity and LSTM both have improvements in the loss of the network.

\begin{table}[h]
\centering
\caption{Network Loss Comparisons}
\label{table_loss}
\begin{tabular}{|c|c|c|}
\hline
\rule{0pt}{9pt}
Method &  Val. Loss ($\times10^{-2}$) & Test Loss ($\times10^{-2}$)\\
\hline
Ours & \textbf{5.59} & \textbf{6.25}\\
\hline
Ours w/o $\omega$ & 6.03 & 7.42\\
\hline
Ours w/o LSTM & 6.27 & 6.56\\
\hline
\end{tabular}
\end{table}

\section{Conclusions}
We present a costmap prediction system that uses a learning method to identify the interactions between the robot and different terrains for autonomous navigation in off-road environments. We introduce a novel traversability cost labelling in the consideration of IMU data and robot risk. Our method incorporates semantic and geometric information of the surface and the robot's velocity as inputs, then outputs a continuous traversability costmap. We demonstrate that our costmaps ensure safe and stable navigation for complex off-road scenarios in comparison of previous work. In the future, hardware experiments will be conducted to validate the system on the robot platform in the real world and adaptation for other vehicles like legged robots would be taken into consideration.

\section*{Acknowledgement}
This work was supported in part by the National Natural Science Foundation of China (NSFC) under Grant No.62273229 and smart city beidou spatial-temporal digital base construction and application industrialization (HCXBCY-2023-020).



\begin{thebibliography}{00}
%

\bibitem{c1} Zürn J, Burgard W, Valada A. Self-supervised visual terrain classification from unsupervised acoustic feature learning. IEEE Transactions on Robotics, 2020, 37(2): 466-481.
\bibitem{c2} Vulpi F, Milella A, Marani R, et al. Recurrent and convolutional neural networks for deep terrain classification by autonomous robots. Journal of Terramechanics, 2021, 96: 119-131.
\bibitem{tc2} Maturana D, Chou P W, Uenoyama M, et al. Real-time semantic mapping for autonomous off-road navigation. Field and Service Robotics: Results of the 11th International Conference. Springer International Publishing, 2018: 335-350.
\bibitem{height1} Meng X, Cao Z, Liang S, et al. A terrain description method for traversability analysis based on elevation grid map. International Journal of Advanced Robotic Systems, 2018, 15: 1–12.
\bibitem{height2} Fankhauser P, Bloesch M, Hutter M. Probabilistic terrain mapping for mobile robots with uncertain localization. IEEE Robotics and Automation Letters, 2018, 3(4): 3019-3026.
\bibitem{app2} Hosseinpoor S, Torresen J, Mantelli M, et al. Traversability analysis by semantic terrain segmentation for mobile robots. 2021 IEEE 17th international conference on automation science and engineering (CASE). IEEE, 2021: 1407-1413.
\bibitem{s1} Dabbiru L, Sharma S, Goodin C, et al. Traversability mapping in off-road environment using semantic segmentation. Autonomous Systems: Sensors, Processing, and Security for Vehicles and Infrastructure 2021. SPIE, 2021, 11748: 78-83.
\bibitem{lc3} Seo J, Sim S, Shim I. Learning Off-Road Terrain Traversability with Self-Supervisions Only. IEEE Robotics and Automation Letters, 2023.
\bibitem{lc4} Frey J, Mattamala M, Chebrolu N, et al. Fast Traversability Estimation for Wild Visual Navigation. arXiv preprint arXiv:2305.08510, 2023.
\bibitem{360} Wu Q, Xu X, Chen X, et al. 360-VIO: A Robust Visual–Inertial Odometry Using a 360° Camera. IEEE Transactions on Industrial Electronics, 2024, 71(9): 11136-11145.
\bibitem{hc2} Guan, Z. He, R. Song, D. Manocha and L. Zhang, TNS: Terrain traversability mapping and navigation system for autonomous excavators. Proceedings of Robotics: Science and Systems, 2022.
\bibitem{hc3} Leung T H Y, Ignatyev D, Zolotas A. Hybrid terrain traversability analysis in off-road environments. 2022 8th International Conference on Automation, Robotics and Applications (ICARA). IEEE, 2022: 50-56.
\bibitem{lc1} Fan D D, Agha-Mohammadi A A, Theodorou E A. Learning risk-aware costmaps for traversability in challenging environments. IEEE robotics and automation letters, 2021, 7(1): 279-286.
\bibitem{lc2} Cai X, Everett M, Fink J, et al. Risk-aware off-road navigation via a learned speed distribution map. 2022 IEEE/RSJ International Conference on Intelligent Robots and Systems (IROS). IEEE, 2022: 2931-2937.
\bibitem{imu1} Sathyamoorthy A J, Weerakoon K, Guan T, et al. Terrapn: Unstructured terrain navigation using online self-supervised learning. 2022 IEEE/RSJ International Conference on Intelligent Robots and Systems (IROS). IEEE, 2022: 7197-7204.
\bibitem{imu2} Waibel G G, Löw T, Nass M, et al. How rough is the path? Terrain traversability estimation for local and global path planning. IEEE Transactions on Intelligent Transportation Systems, 2022, 23(9): 16462-16473.
\bibitem{imu3} Seo J, Kim T, Kwak K, et al. Scate: A scalable framework for self-supervised traversability estimation in unstructured environments. IEEE Robotics and Automation Letters, 2023, 8(2): 888-895.
\bibitem{imu4} Yao X, Zhang J, Oh J. Rca: Ride comfort-aware visual navigation via self-supervised learning. 2022 IEEE/RSJ International Conference on Intelligent Robots and Systems (IROS). IEEE, 2022: 7847-7852.
\bibitem{imu5} Castro M G, Triest S, Wang W, et al. How does it feel? self-supervised costmap learning for off-road vehicle traversability. 2023 IEEE International Conference on Robotics and Automation (ICRA). IEEE, 2023: 931-938.
\bibitem{imu_gaussian} Aranburu A. IMU Data Processing to Recognize Activities of Daily Living with a Smart Headset. University of California, Santa Cruz, 2018.
\bibitem{imu_gaussian_2} Nirmal K, Sreejith A G, Mathew J, et al. Noise modeling and analysis of an IMU-based attitude sensor: improvement of performance by filtering and sensor fusion. Advances in optical and mechanical technologies for telescopes and instrumentation II. SPIE, 2016, 9912: 2138-2147.
\bibitem{risk} Majumdar A, Pavone M. How should a robot assess risk? towards an axiomatic theory of risk in robotics. Robotics Research: The 18th International Symposium ISRR. Springer International Publishing, 2020: 75-84.
\bibitem{fast-scnn} Poudel R P K, Liwicki S, Cipolla R. Fast-scnn: Fast semantic segmentation network. arXiv preprint arXiv:1902.04502, 2019.
\bibitem{resnet} He K, Zhang X, Ren S, et al. Deep residual learning for image recognition. Proceedings of the IEEE conference on computer vision and pattern recognition (CVPR), 2016: 770-778.
\bibitem{lstm} Graves A, Graves A. Long short-term memory. Supervised sequence labelling with recurrent neural networks, 2012: 37-45.
\bibitem{gazebo} Sánchez M, Morales J, Martínez J L, et al. Automatically annotated dataset of a ground mobile robot in natural environments via gazebo simulations. Sensors, 2022, 22(15): 5599.
\bibitem{falco} Zhang J, Hu C, Chadha R G, et al. Falco: Fast likelihood‐based collision avoidance with extension to human‐guided navigation. Journal of Field Robotics, 2020, 37(8): 1300-1313.



\end{thebibliography}
\end{document}